\documentclass[a4paper,fleqn]{cas-sc}

\usepackage[authoryear]{natbib}
\usepackage{tablefootnote}
\usepackage{lineno}
\usepackage{tcolorbox}
\usepackage{subcaption}
\usepackage{tablefootnote}
\usepackage{algorithm}
\usepackage{algpseudocode}

\def\tsc#1{\csdef{#1}{\textsc{\lowercase{#1}}\xspace}}
\tsc{WGM}
\tsc{QE}

\ExplSyntaxOn
\cs_set:Npn \__first_footerline:
  {
    \group_begin:
    \small
    \sffamily
    \ifnum\theblind>0\relax
    \else
      \__short_authors:
    \fi
    \group_end:
  }
\ExplSyntaxOff

\begin{document}
\let\WriteBookmarks\relax
\def\floatpagepagefraction{1}
\def\textpagefraction{.001}

\shorttitle{}

\shortauthors{Y. Quan}

\title [mode = title]{AutoMARS: Automatic and Scalable Multi-Agent Large Language Model Framework for Supply Chain Risk Assessment}

\author[1]{Yinzhu Quan}[orcid=0009-0003-6119-9009]
\cormark[1]
\ead{yquan9@gatech.edu}

\author[2]{Zefang Liu}[orcid=0000-0003-1947-3249]
\ead{liuzefang@gatech.edu}

\author[1,3]{Frederick Benaben}[orcid=0000-0002-1676-6524]
\ead{frederick.benaben@mines-albi.fr}

\author[1]{Benoit Montreuil}[orcid=0000-0001-9909-9212]
\ead{benoit.montreuil@isye.gatech.edu}

\affiliation[1]{organization={Physical Internet Center, H. Milton Stewart School of Industrial and Systems Engineering, Georgia Institute of Technology},
            city={Atlanta},
            postcode={30332},
            state={GA},
            country={USA}}

\affiliation[2]{organization={School of Computational Science and Engineering, Georgia Institute of Technology},
            city={Atlanta},
            postcode={30332},
            state={GA},
            country={USA}}

\affiliation[3]{organization={Centre Genie Industriel, IMT Mines Albi},
            city={Albi},
            country={France}}

\cortext[1]{Corresponding author}

\begin{abstract}
We propose AutoMARS (\underline{Auto}matic \underline{M}ulti-\underline{A}gent \underline{R}isk A\underline{s}sessment), an agentic framework for automatic and scalable comparative risk assessment in supply chain networks. The framework enables end-to-end execution of analytical workflows by automatically planning tasks, coordinating specialized agents, invoking external analytical tools, and synthesizing domain-specific results into integrated risk evaluations. AutoMARS further supports scalable risk comparison across multiple locations by parallelizing independent analytical tasks. The framework adopts a multi-domain, multi-agent, and multi-tool architecture that integrates heterogeneous risk factors from socioeconomic, disaster, environmental, and transportation domains. Through coordinated agent interactions and tool integration, AutoMARS enables structured, systematic, and interpretable comparative analysis to support supply chain risk-aware location evaluation.
\end{abstract}


\maketitle

\section{Introduction}

Modern supply chains operate in increasingly volatile, uncertain, complex, and ambiguous (VUCA) environments. In such environments, companies increasingly depend on geographically dispersed suppliers, complex logistics networks, and digitally coordinated operations to sustain production and distribution activities. While these configurations improve efficiency and global reach, they also expand the exposure of supply chains to diverse disruption sources. Events such as natural hazards, environmental disruptions, transportation accidents, and socioeconomic vulnerabilities can interrupt material flows, delay logistics operations, and create instability across multiple stages of supply chains. Consequently, managing disruption risk has become a critical challenge for organizations seeking to maintain supply continuity and operational resilience in increasingly uncertain environments~\citep{ivanov2022viable, katsaliaki2022supply}.

To address the above challenge, extensive research has focused on developing analytical frameworks and decision-support methods for assessing and managing risks in interconnected supply chain networks~\citep{montreuil2011toward,montreuil2012physical}. Existing studies have explored various approaches to support risk identification, evaluation, and mitigation in complex supply chain environments, including optimization-based models~\citep{ivanov2018structural}, simulation methods~\citep{cerabona2024physics}, and multi-agent coordination mechanisms~\citep{giannakis2011multi}. These approaches aim to improve the ability of supply chain systems to anticipate disruptions, evaluate potential impacts, and support informed operational and strategic decision-making. More recently, advances in artificial intelligence (AI) and large language models (LLMs) have opened new opportunities for enhancing risk assessment capabilities in complex supply networks. For example, \citet{quan2026leveraging} propose a novel LLM-enhanced framework, MARS, that integrates both structured and unstructured factors to support risk-aware logistic hub site selection.

Despite the advantages of existing approaches, several key limitations remain. First, fully integrated end-to-end automation is still limited, as many multi-agent systems (MAS) require partial human intervention and manual coordination across analytical steps~\citep{ambati2025event}, preventing seamless execution from task decomposition to final assessment. Second, scalability remains a challenge in large-scale risk assessment scenarios~\citep{katsaliaki2022supply}, where sequential interactions with data sources and analytical agents lead to significant computational overhead, highlighting the need for parallel execution. Third, reliance on absolute risk scores can reduce stability when LLM-based agents are involved, as small output variations may affect final results~\citep{aqlan2015fuzzy, babu2023supply}. Comparative evaluation based on relative ordering provides a more robust alternative, improving consistency and interpretability across repeated assessments.

In this paper, we propose AutoMARS (\underline{Auto}matic \underline{M}ulti-\underline{A}gent \underline{R}isk A\underline{s}sessment), an agentic framework designed to support automatic and scalable comparative risk assessment in supply chain networks. The framework enables end-to-end execution of the analytical workflow, allowing a user to submit a high-level request while the system automatically plans and coordinates multiple specialized agents to invoke external tools, perform domain-specific analyses, and synthesize the results into an integrated risk evaluation. By integrating multi-domain information, coordinating multiple agents, and invoking diverse analytical tools, AutoMARS provides a unified framework for complex risk analysis tasks. AutoMARS further supports scalable analysis across multiple locations by enabling parallel execution of independent analytical tasks. In addition, the framework emphasizes comparative risk evaluation across locations, enabling more consistent and reliable decision-making by focusing on relative risk ordering rather than absolute scores. This comparative perspective provides a more robust and interpretable basis for decision-making in supply chain location evaluation. Risk-conscious logistic location selection is essential for facility location and network design decisions, including hub siting and infrastructure investment. Overall, this work aims to develop an automatic, scalable, and reliable framework for comparative supply chain risk assessment across multiple locations.

In this study, we aim to address the following research questions (RQs):
\begin{itemize}
    \item \textbf{RQ1:} How to enable end-to-end automation in supply chain risk assessment?
    \item \textbf{RQ2:} How to scale supply chain risk assessment while mitigating the computational complexity introduced by multi-agent coordination?
    \item \textbf{RQ3:} How to conduct comparative risk assessment across multiple locations?
\end{itemize}

Answering these research questions, our main contributions in this paper are as follows:
\begin{enumerate}
    \item \textbf{End-to-end automatic framework for supply chain risk assessment:} We propose AutoMARS, an automatic multi-agent framework that enables end-to-end execution of supply chain risk assessment tasks. Given a high-level user request, the framework automatically plans and decomposes the analytical workflow, coordinates specialized agents, invokes relevant external tools, and synthesizes domain-specific results into a unified risk evaluation.
    \item \textbf{Scalable risk comparison through parallel execution:} The proposed framework enables scalable comparative risk assessment across multiple locations by executing independent analytical tasks in parallel. This design significantly reduces computational time when evaluating many locations and supports efficient risk comparison in large-scale supply chain decision contexts.
    \item \textbf{Multi-domain risk analysis through multi-agent multi-tool integration:} The framework integrates heterogeneous risk factors across multiple domains, including socioeconomic, disaster, environmental, and transportation domains, by coordinating specialized agents and invoking diverse external analytical tools. This multi-domain, multi-agent, and multi-tool architecture enables comprehensive risk analysis by combining information from multiple sources and analytical perspectives.
\end{enumerate}

The rest of this paper is organized as follows. Section~\ref{sec:related_work} reviews related work on supply chain risk management, LLM-based multi-agent orchestration, and LLM-as-a-judge evaluation. Section~\ref{sec:problem_definition} defines the problem of comparative supply chain risk assessment. Section~\ref{sec:methodology} presents the proposed AutoMARS framework, including system architecture, agent design, and execution strategies. Section~\ref{sec:experiments} reports experimental results and evaluates the effectiveness, stability, and scalability of the framework. Section~\ref{sec:conlusion} concludes the paper and discusses future research directions.

\section{Related Work}
\label{sec:related_work}

In this section, we review prior research relevant to this study. We first review existing studies on automation in supply chain risk management (SCRM). Next, we discuss recent developments in large language model (LLM)-based multi-agent orchestration frameworks. Finally, we discuss the emerging paradigm of LLM-as-a-Judge for evaluating multi-agent reasoning and outcomes.

\subsection{Automation in Supply Chain Risk Management}

SCRM involves continuous tasks such as risk identification, assessment, monitoring, and response~\citep{fan2017information}, which are difficult to sustain under rapidly changing conditions. In practice, risk assessment often relies on expert judgment and static analyses, limiting scalability across regions and time~\citep{sodhi2012researchers, ho2015supply}. Meanwhile, risk information arises from heterogeneous, high-frequency sources, creating the need for automated processing and timely updates~\citep{chu2020global}. This motivates artificial intelligence (AI)-driven automation in SCRM to improve efficiency, consistency, and scalability~\citep{ganesh2022future, ziari2025integrated, chowdhury2026navigating}.

Existing automation approaches support the representation and modeling of temporal events from text, enabling standardized analysis of evolving risks~\citep{liu2024tpp, liu2025retrieval, liu2025temporal}. They also integrate multi-source data for risk assessment and decision support in logistics contexts~\citep{zhao2024optimizing, quan2025leveraging}. More broadly, generative AI and automated decision systems enhance analytical capabilities and responsiveness~\citep{oger2018towards, sharma2022future, jackson2024generative}, with applications in areas such as inventory management demonstrating efficiency gains~\citep{saleem2020automated}.

Despite these advances, automation in SCRM remains fragmented, with most approaches focusing on isolated functions rather than end-to-end workflows~\citep{das2025supply} and offering limited support for systematic risk comparison across regions~\citep{katsaliaki2022supply}. Simulation-based and digital twin systems enable automated monitoring but rely on predefined models, limiting adaptability~\citep{ivanov2021digital}. Evidence from automated multi-agent systems (MAS) in other domains further suggests the need for more integrated frameworks~\citep{liu2025autobnb}. Overall, automation enhances efficiency, consistency, and responsiveness in SCRM by enabling data integration and supporting coordinated risk assessment.

\subsection{Automated LLM-Based Multi-Agent Orchestration}

Language interaction enables flexible coordination among decision-making entities in complex systems. Traditional MAS rely on predefined protocols, limiting adaptability to dynamic contexts and heterogeneous information. In contrast, LLM-based agents use natural language to coordinate roles, enabling orchestration through conversational workflows, role assignment, and task decomposition~\citep{li2023camel, wu2024autogen, hong2023metagpt, guo2024large}.

Existing work spans domain-specific applications and general-purpose frameworks. Domain-oriented systems use LLM-driven agents coordinated through automated workflows for tasks such as inventory management and customer-facing processes~\citep{quan2024invagent, quan2025crmagent}. General approaches orchestrate multiple agents or tools via task decomposition to solve complex problems end to end~\citep{shen2023hugginggpt, schick2023toolformer}.

In practice, multi-agent coordination is enabled at multiple levels. Frameworks such as LangChain\footnote{\url{https://www.langchain.com/}} and LangGraph\footnote{\url{https://www.langchain.com/langgraph}} support structured orchestration through pipeline- or graph-based workflows, while protocols such as Model Context Protocol\footnote{\url{https://www.anthropic.com/news/model-context-protocol}} (MCP) and Agent2Agent Protocol\footnote{\url{https://developers.googleblog.com/en/a2a-a-new-era-of-agent-interoperability/}} (A2A) standardize agent interoperability and tool integration. Building on these foundations, systems such as BMAD\footnote{\url{https://github.com/bmad-code-org/BMAD-METHOD}}, GSD\footnote{\url{https://github.com/gsd-build/gsd-2}}, and OpenCLAW\footnote{\url{https://openclaw.ai/}} further demonstrate specification-driven workflows and structured multi-agent orchestration.

Despite recent progress, LLM-based multi-agent orchestration faces challenges in stability and evaluation~\citep{li2024survey, guo2024large, tran2025multi}. Ensuring alignment, safety, and controllability in coordinated workflows remains critical for reliable deployment~\citep{bai2022constitutional}. Overall, automated multi-agent orchestration enables flexible coordination, task decomposition, and scalable reasoning across agents and tools, but further advances are required to ensure robustness, stability, and well-governed execution in real-world settings.

\subsection{LLM-as-a-Judge in Multi-Agent Collaboration}

Evaluating complex reasoning and collaborative outcomes remains a key challenge in LLM-driven MAS. Traditional approaches rely on handcrafted metrics or human annotation, which do not scale well to multi-step interactive settings. Recent work uses LLMs as evaluators to assess outputs, reasoning processes, and task performance, enabling flexible evaluation~\citep{bai2022constitutional, madaan2023self}. LLM-based judges further support preference comparison and feedback generation and have been proposed as scalable alternatives to human evaluation in multi-agent settings~\citep{li2024llms, li2025generation, thakur2025judging}. However, using the same LLM for both generation and evaluation may introduce shared biases, although empirical studies show reasonable alignment with human judgment.

LLM-as-a-Judge has been applied to evaluate agent behavior, reasoning quality, and task outcomes in interactive environments. Early studies demonstrate automatic assessment of outputs and reasoning traces~\citep{yue2023automatic, zhang2023wider}, while later work extends evaluation to multi-step interactions, tool usage, and task completion~\citep{xu2023towards, chiang2024chatbot}. Recent benchmarks formalize this paradigm by comparing agent outputs across diverse tasks and interaction patterns~\citep{tan2024judgebench}. Domain-specific applications further evaluate sequential reasoning and agent performance in economic and web-based settings~\citep{quan2024econlogicqa, liu2025econwebarena}.

Despite its promise, LLM-as-a-Judge faces limitations in reliability and standardization. Prior work shows sensitivity to context, inconsistent judgments~\citep{chen2025multi, szymanski2025limitations}, and risks of bias amplification and feedback effects~\citep{kiela2021dynabench, saunders2022self}. The lack of standardized protocols and concerns about transparency and alignment with human judgment further motivate more robust and holistic evaluation frameworks~\citep{liang2022holistic, chen2024humans, bavaresco2025llms}. To mitigate these issues, recent studies explore strategies such as multi-judge aggregation, cross-model evaluation, structured criteria, and hybrid human-AI evaluation to improve robustness~\citep{gu2024survey}. Overall, LLM-as-a-Judge offers a flexible paradigm for evaluating complex multi-agent reasoning, while its reliability can benefit from human oversight.

\section{Problem Definition}
\label{sec:problem_definition}

Logistic location selection is a critical decision in supply chain networks, where evaluating and comparing risk levels across different geographic regions plays an important role in ensuring operational stability and resilience. Regions may differ significantly in terms of socioeconomic conditions, disaster exposure, environmental quality, and transportation safety, all of which can affect the reliability of supply chain operations. However, previous risk assessment approaches often require manual coordination across multiple analytical steps, provide limited capability to integrate heterogeneous risk information from multiple domains, and require substantial computational time when evaluating risks across many regions. These limitations restrict the ability to conduct efficient and comprehensive comparative risk assessment across regions, particularly when a large number of locations must be evaluated. As supply chain decision-making increasingly requires integrating heterogeneous risk information and comparing risks across multiple regions, a key challenge is to develop frameworks that can automatically coordinate multi-domain analyses, support parallel task execution, and efficiently synthesize risk indicators from diverse information sources. This highlights the need for intelligent systems capable of planning tasks, coordinating multiple analytical components, and synthesizing diverse forms of risk information to enable comprehensive and adaptive location risk evaluation.

\textbf{Problem Description:}
Consider a set of candidate locations within a supply chain network. For each location, risk information can be obtained from multiple domains. To analyze these heterogeneous risk indicators, a set of specialized domain agents is available, where each agent is designed to handle a specific analytical task. Each agent can access a collection of external tools that provide domain-specific data retrieval, processing, and analysis capabilities. The objective is to enable Manager Agent to automatically plan and decompose the overall risk assessment workflow, coordinate the execution of Domain Agents and their associated tools, and integrate multi-domain information into a unified risk evaluation for each location. Reflection Agent evaluates the synthesized results and provides feedback, allowing Manager Agent to revise the analysis until a consistent comparative risk assessment is obtained. To further improve scalability, independent analytical tasks within the workflow can be executed in parallel.

In summary, the problem of comparative supply chain risk assessment requires a framework capable of coordinating domain-specific analyses and integrating heterogeneous risk information across multiple locations. To address this challenge, we propose an automatic multi-agent framework that supports dynamic task orchestration, parallel execution, and iterative reflection for scalable and adaptive risk evaluation.

\section{Methodology}
\label{sec:methodology}

This section presents the methodology of AutoMARS, including its overall framework, agent architecture, and efficient execution strategies. We first describe the system architecture and workflow for structured multi-agent coordination. Next, we detail the design and interactions of specialized agents for cross-domain risk analysis. Finally, we introduce execution strategies that improve efficiency, including parallelism, caching mechanisms, and a sorting algorithm.

\subsection{Framework}

\begin{figure}[pos=h]
    \centering
    \includegraphics[width=0.9\linewidth]{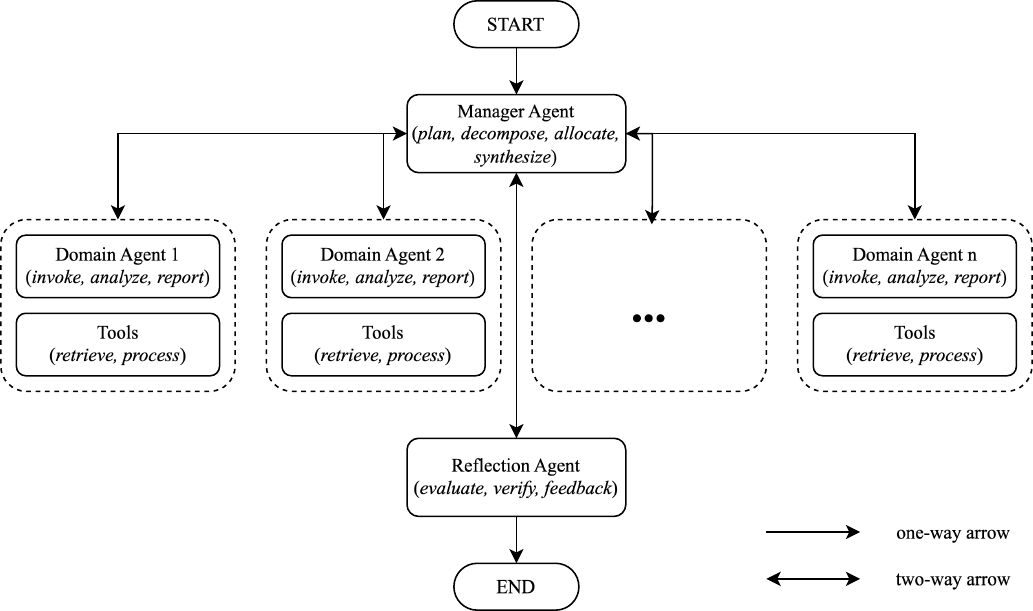}
    \caption{Framework of AutoMARS for supply chain risk assessment.}
    \label{fig:inner-loop}
\end{figure}

AutoMARS is an automatic multi-agent framework for comparative risk assessment across multiple locations in supply chain networks. To enable end-to-end automation (RQ1), it supports coordinated planning, tool invocation, and cross-domain synthesis within a unified workflow. As illustrated in Figure~\ref{fig:inner-loop}, the framework adopts a modular architecture consisting of a Manager Agent, multiple Domain Agents with associated external tools, and a Reflection Agent. Manager Agent plans and orchestrates the overall assessment workflow, decomposing the task into domain-specific analyses and coordinating Domain Agents, which invoke external tools to retrieve and process heterogeneous risk information. The results are then integrated into a comparative risk evaluation. Reflection Agent subsequently evaluates the integrated results to verify the completeness and consistency of the reasoning process. If deficiencies are identified, feedback is returned to Manager Agent, which revises the assessment accordingly. This iterative interaction continues until a stable and reliable comparative risk assessment is obtained. The detailed workflow of AutoMARS framework is formally summarized in Algorithm~\ref{alg:multi_agent_workflow}.

\begin{algorithm}[t]
\caption{Parallel Multi-Agent Workflow}
\label{alg:multi_agent_workflow}

\begin{algorithmic}[1]

\State \textbf{Input:} User query $q$, manager agent $M$, reflection agent $R$, domain worker agents $\{W_1, W_2, \dots, W_n\}$, tool sets $\{\mathcal{T}_1, \mathcal{T}_2, \dots, \mathcal{T}_n\}$
\State \textbf{Output:} Final response

\State Initialize shared state
$\mathcal{S} \leftarrow
\{\text{messages}: [q],\ \text{reports}: \emptyset \}$

\Function{WorkerSubgraph}{$W_i,\mathcal{T}_i,\mathcal{S}$}

    \State Initialize private worker state $\mathcal{S}_i$

    \Repeat

        \State Generate worker response
        $m_i \leftarrow W_i(\mathcal{S}_i)$

        \If{$m_i$ contains tool calls}

            \State Execute requested tools from $\mathcal{T}_i$

            \State Append tool outputs to $\mathcal{S}_i$

        \EndIf

    \Until{$m_i$ contains no further tool calls}

    \State Set worker report $r_i \leftarrow m_i$

    \State \Return $r_i$

\EndFunction

\Repeat

    \State Generate manager response
    $m_M \leftarrow M(\mathcal{S})$

    \State Extract selected worker set $\mathcal{W}^{*}$ from $m_M$

    \If{$\mathcal{W}^{*}$ is not empty}

        \State Execute selected workers in parallel:
        $\{r_i\}_{W_i \in \mathcal{W}^{*}}
        \leftarrow
        \Call{WorkerSubgraph}{W_i, \mathcal{T}_i, \mathcal{S}}$

        \State Merge worker reports $\{r_i\}$ into $\mathcal{S}$

    \EndIf

    \State Generate reflection response
    $m_R \leftarrow R(\mathcal{S})$

\Until{$m_R$ accepts the assessment}

\State \Return final response
\end{algorithmic}
\end{algorithm}

\subsection{Agents}
\label{sec:agents}

This section describes agents in AutoMARS and their roles in the overall workflow. The system includes Manager Agent for coordination and integration, Domain Agents for specialized analysis, and Reflection Agent for validating the final results.

\subsubsection{Manager Agent}

Manager Agent serves as the central coordinator of the AutoMARS framework and is responsible for planning, orchestrating, and integrating the overall risk assessment workflow. Given a high-level user query, Manager Agent generates an initial analysis plan that determines which Domain Agents should be activated and how the analytical process should proceed. The behavior of Manager Agent is guided by a structured system prompt that specifies its planning, execution, and integration responsibilities. A simplified version of the prompt is summarized (see ``Prompt for Manager Agent (Condensed Version)'').

\begin{tcolorbox}[title=\textbf{Prompt for Manager Agent (Condensed Version)}, colback=gray!5, colframe=gray!40!black]

{\small
You are Manager Agent responsible for orchestrating multiple Domain Agents to perform a comparative risk assessment between Location A and Location B. Your task is to plan, execute, and integrate the analytical workflow. Domain Agents provide domain-specific evidence, while you control the workflow and decision logic.\\

\textbf{Stage 1: Initial Planning.}
At the beginning of the workflow, Manager Agent constructs an initial analysis plan before consulting any Domain Agent. The plan specifies the domains to be analyzed, identifies which Domain Agents should be activated, and determines which analyses can be executed in parallel. It also defines the role of each Domain Agent and outlines how evidence from multiple domains will be integrated to support the final comparative risk judgment.\\

\textbf{Stage 2: Evidence-Guided Execution.}
Manager Agent coordinates the execution of Domain Agents and collects domain-specific evidence as analyses are completed. During this process, Manager Agent summarizes the obtained results, evaluates whether the current plan remains appropriate, and revises the execution strategy when new evidence suggests that adjustments are necessary.\\

\textbf{Stage 3: Reflection Feedback Handling.}
After synthesizing the comparative risk assessment, Manager Agent submits the result to Reflection Agent for validation. If Reflection Agent identifies deficiencies, Manager Agent revises the comparative assessment according to the feedback while preserving previously generated domain evidence. If additional domain analyses are required, updated evidence is incorporated into the synthesis. This revision loop continues until Reflection Agent accepts the final assessment.
}

\end{tcolorbox}

Based on the above prompt specification, Manager Agent operates in a structured three-stage process. In the initial planning stage, it decomposes the comparative risk assessment task into multiple domain-specific analyses and determines the execution strategy, including the selection of Domain Agents and opportunities for parallel execution. In the evidence-guided execution stage, it coordinates Domain Agents and incrementally collects domain-level evidence while evaluating and updating the execution plan as new information becomes available. In the reflection feedback stage, it synthesizes the multi-domain evidence into a unified comparative risk assessment and submits the result to Reflection Agent for validation. If deficiencies are identified, the assessment is revised by incorporating updated evidence until a consistent and reliable final judgment is obtained.

\subsubsection{Domain Agents}

AutoMARS incorporates multiple Domain Agents to capture heterogeneous risk factors across different dimensions of supply chain environments. Specifically, the framework includes Census Agent, Disaster Agent, Environment Agent, and Traffic Agent, which respectively focus on socioeconomic conditions, hazard exposure, environmental quality, and transportation safety. Each Domain Agent is equipped with a set of specialized external tools for retrieving and analyzing domain-specific data. Tools used by different Domain Agents, along with their corresponding data source types and functional roles, are summarized in Table~\ref{tab:tools}.

\begin{table}[pos=h]
    \centering
    \caption{Tools used in AutoMARS, including their associated agents, data source types, and functional descriptions.}
    \label{tab:tools}
    \begin{tabular}{llll}
        \toprule
        \textbf{Tool} & \textbf{Agent} & \textbf{Type} & \textbf{Description} \\
        \midrule

        Labor Exposure Tool\tablefootnote{{\url{https://www.census.gov/programs-surveys/ces/data/restricted-use-data/lehd-data.html}}} & Census & API & Extract labor composition indicators \\
        Workforce Dynamics Tool\tablefootnote{{\url{https://www.census.gov/data/developers/data-sets/qwi.html}}} & Census & API & Capture employment flow and turnover indicators \\
        Socioeconomic Profile Tool\tablefootnote{{\url{https://www.census.gov/programs-surveys/acs.html}}} & Census & API & Retrieve core socioeconomic indicators \\
        Community Resilience Tool\tablefootnote{{\url{https://www.census.gov/data/developers/data-sets/community-resilience-estimates.html}}} & Census & API & Characterize population risk factor distributions \\
        Industry Scale Tool\tablefootnote{{\url{https://www.census.gov/data/developers/data-sets/economic-census.html}}} & Census & API & Quantify industry-level economic scale indicators \\

        Disaster Declaration Tool\tablefootnote{{\url{https://www.fema.gov/openfema-data-page/disaster-declarations-summaries-v2}}} & Disaster & API & Identify federal disaster declaration events \\
        Multi-Hazard Risk Index Tool\tablefootnote{{\url{https://hazards.fema.gov/nri/data-resources}}} & Disaster & API & Provide composite multi-hazard risk scores \\
        Climate Tool\tablefootnote{{\url{https://www.ncdc.noaa.gov/cdo-web/webservices/v2}}} & Disaster & API & Query historical climate observations \\
        Severe Weather Tool\tablefootnote{{\url{https://www.ncdc.noaa.gov/stormevents/}}} & Disaster & API & Aggregate hazard event frequency and impact indicators \\

        Air Quality Tool\tablefootnote{{\url{https://www.epa.gov/aqs}}} & Environment & API & Aggregate air quality exposure indicators \\
        Environmental Facility Tool\tablefootnote{{\url{https://www.epa.gov/frs}}} & Environment & API & Identify regulated environmental facilities \\
        Social Vulnerability Index Tool\tablefootnote{{\url{https://www.atsdr.cdc.gov/place-health/php/svi/svi-data-documentation-download.html}}} & Environment & API & Assess relative social vulnerability levels \\

        Traffic Accident Tool\tablefootnote{{\url{https://smoosavi.org/datasets/us_accidents}}} & Traffic & Database & Aggregate traffic accident frequency and severity indicators \\

        \bottomrule
    \end{tabular}
\end{table}

\paragraph{\textbf{Census Agent.}} Census Agent is responsible for characterizing the socioeconomic and labor conditions of a region, which provide evidence on human and economic exposure, population vulnerability, and underlying socioeconomic conditions that influence how risks manifest across locations. This agent focuses on indicators describing population composition, labor market conditions, economic activity, and community characteristics.

To regulate the behavior of Census Agent, a structured prompt is designed to define its analytical scope, data retrieval strategy, and output constraints. The prompt ensures that the agent identifies relevant socioeconomic indicators, retrieves data efficiently using appropriate tools, and produces consistent, evidence-based comparative outputs without introducing subjective judgments or workflow decisions. A simplified version of the prompt is presented below for illustration purposes, using Census Agent as a representative example (see “Prompt for Census Agent (Condensed Version)”).

\begin{tcolorbox}[title=\textbf{Prompt for Census Agent (Condensed Version)}, colback=gray!5, colframe=gray!40!black]

{\small

You are Census Agent responsible for analyzing socioeconomic and demographic conditions for comparative risk assessment between Location A and Location B. Your role is to retrieve and compare indicators related to population structure, labor force characteristics, economic activity, and mobility patterns, in order to characterize differences in human and economic exposure across locations.\\

Census Agent must identify required indicators in advance and retrieve relevant datasets efficiently, prioritizing recent and complete summary data. The analysis should be strictly evidence-based, clearly distinguish the two locations, and produce structured outputs suitable for downstream comparative risk evaluation. The agent must not provide recommendations or influence workflow decisions.

}

\end{tcolorbox}

\paragraph{\textbf{Disaster Agent.}} Disaster Agent is responsible for characterizing historical hazard exposure and disaster-related risk at the regional level, which provides essential context for assessing the frequency, severity, and consequences of disruptive events. This agent focuses on hazard occurrence, disaster declarations, climate conditions, and multi-hazard risk indicators that capture both acute shocks and longer-term environmental threats.

\paragraph{\textbf{Environment Agent.}} Environment Agent is responsible for characterizing environmental exposures and infrastructure-related conditions that influence regional risk and vulnerability. This agent focuses on ambient environmental quality, regulated facility presence, and social vulnerability patterns that reflect chronic exposure and underlying environmental burdens.

\paragraph{\textbf{Traffic Agent.}} Traffic Agent is responsible for characterizing transportation-related safety risk and infrastructure exposure at the regional level. This agent focuses on the frequency and severity of traffic accidents as indicators of mobility-related risk and infrastructure vulnerability.

\subsubsection{Reflection Agent}

Reflection Agent serves as the final validation component of AutoMARS framework and is responsible for evaluating the completeness and consistency of the integrated risk assessment results. Given the outputs synthesized by Manager Agent from all Domain Agents, Reflection Agent determines whether the assessment is sufficient for decision support by verifying cross-domain evidence coverage, logical consistency, and the clarity of the final comparative conclusion. The behavior of Reflection Agent is guided by a structured system prompt that specifies its validation criteria and decision routing logic. For readability, a simplified version of the prompt is summarized (see ``Prompt for Reflection Agent (Condensed Version)'').

\begin{tcolorbox}[title=\textbf{Prompt for Reflection Agent (Condensed Version)}, colback=gray!5, colframe=gray!40!black]

{\small

You are the Reflection Agent. Your task is to validate the final comparative risk assessment between Location A and Location B.
You do not perform analysis, planning, or refinement; you only evaluate whether the result is sufficient for decision support.\\

The assessment is acceptable if:\\
1. All required domains are meaningfully referenced\\
2. A clear higher-risk location is explicitly identified\\
3. The conclusion is supported by consistent, domain-specific evidence\\

If the assessment is acceptable:\\
- Produce a concise final determination\\
- Summarize cross-domain evidence\\
- Output a structured result and terminate the workflow\\

If the assessment is not acceptable:\\
- Identify missing or inconsistent components\\
- Specify which domains require revision\\
- Route back to the Manager Agent

}

\end{tcolorbox}

Based on the above prompt specification, Reflection Agent determines whether the integrated assessment is sufficiently complete and logically consistent for final decision support. The assessment is accepted when all required domains are meaningfully incorporated, cross-domain evidence supports a consistent comparative conclusion, and a clear higher-risk location is explicitly identified. In this case, Reflection Agent produces the final structured output and terminates the workflow. Otherwise, if missing domain evidence, unclear conclusions, or logical inconsistencies are detected, the agent identifies the problematic components and routes the workflow back to Manager Agent for further revision and coordination.

\subsection{Efficient Execution}

This section presents the key mechanisms that enable efficient execution in AutoMARS, including parallel execution for reducing latency, caching mechanisms for avoiding redundant computation, and a comparison-based sorting algorithm for scalable ranking across locations.

\subsubsection{Parallel Execution}

AutoMARS adopts a hierarchical parallel execution strategy to support efficient and scalable execution of multiple independent tasks (RQ2). The system is implemented as a graph-based execution model in which Manager Agent dynamically dispatches multiple Domain Agents as independent worker subgraphs that can execute concurrently. Within each subgraph, tool calls are asynchronously executed through a dedicated tool execution module to support parallel data retrieval from heterogeneous external sources. Intermediate outputs are stored in structured shared states indexed by unique keys and merged through synchronized associative updates before being passed back to Manager Agent for integration. By enabling parallel processing at both inter-agent and intra-agent levels, the system minimizes idle waiting time while maintaining modularity and supporting scalable extension to additional domains and data sources. This fan-out and fan-in execution pattern enables efficient multi-level parallel processing while ensuring deterministic aggregation of multi-domain evidence.

\subsubsection{Caching Mechanism}

AutoMARS incorporates a caching mechanism to reduce redundant computation and improve efficiency in repeated workflows involving external data retrieval and large language model (LLM) inference. The caching mechanism is implemented through multiple cache layers tailored to different system components. For external tools, the system maintains a disk-based cache in which final structured outputs are stored as persistent JSON files indexed by deterministic identifiers constructed from input locations. Upon receiving a request, the system performs a lookup using the corresponding identifier; matching entries are directly reused, while missing entries trigger new execution and storage for future reuse. For LLM calls, caching is handled separately through a key-based mechanism in which prompts and contextual inputs are normalized and serialized into a canonical representation before generating a hash-based cache key. Upon each invocation, the system computes the corresponding key and checks for its existence in the cache; matching keys enable direct reuse of previous responses and avoid repeated model inference. Together, these mechanisms enable efficient and consistent reuse of both external data retrieval and repeated LLM computation.

\subsubsection{Sorting Algorithm}

To rank locations by overall risk, AutoMARS adopts a comparison-based sorting procedure in which ordering decisions are determined through pairwise comparative risk assessments rather than predefined scalar scores. For each location pair, the system invokes AutoMARS graph to generate a structured comparison result identifying the higher-risk location, which is then mapped to an ordering relation used by the sorting algorithm. This design derives ranking decisions directly from cross-domain evidence synthesized during multi-agent analysis, avoiding fixed global weighting schemes while supporting interpretable comparative risk ranking. The underlying sorting algorithm follows a stable and adaptive strategy based on Timsort\footnote{\url{https://en.wikipedia.org/wiki/Timsort}}, which combines merge-based and insertion-based techniques to efficiently handle partially ordered data, achieving a time complexity of $O(n \log n)$ in the average and worst cases and $O(n)$ in the best case~\citep{auger2018worst}.

\section{Experiments}
\label{sec:experiments}

This section evaluates AutoMARS in terms of effectiveness, stability, and scalability for comparative risk assessment. We first introduce the experimental setup, followed by pairwise experimental results to demonstrate the workflow and comparative performance. Finally, we present ranking experimental results, analyzing the consistency and robustness of the framework in multi-location ranking scenarios.

\subsection{Experimental Setup}

\textbf{Geographical Scope.} We conduct experiments on counties within Atlanta metropolitan area at multiple spatial scales, including Atlanta Urban Core, Atlanta Regional Commission (ARC) region, and Metro Atlanta. Atlanta Urban Core is defined as a five-county subset representing the most densely populated and economically active areas. ARC region corresponds to an 11-county regional planning area, while Metro Atlanta is defined as Atlanta-Sandy Springs-Alpharetta Metropolitan Statistical Area, consisting of 29 counties. This multi-scale geographical design enables systematic evaluation of the proposed framework across different levels of spatial aggregation.

\textbf{Model Configuration.} All agents in AutoMARS are implemented as large language model (LLM)-driven modules using GPT-5\footnote{\url{https://developers.openai.com/api/docs/models/gpt-5}} as the underlying reasoning engine. The system uses a temperature value of 1 during response generation. All interactions with the model are conducted through API-based invocation with standardized prompts and input formats.

\textbf{Framework Implementation.} AutoMARS is implemented using LangGraph to support graph-based multi-agent orchestration. LangGraph provides the workflow structure for coordinating Manager Agent, Domain Agents, and Reflection Agent, while maintaining shared state across execution steps. It also supports parallel execution of independent agent branches and iterative routing for reflection-based revision.

\subsection{Pairwise Experimental Results}

This subsection presents pairwise comparative experiments to evaluate the effectiveness of AutoMARS in two-location risk assessment scenarios. We first provide a detailed case study to illustrate the end-to-end workflow and cross-domain reasoning process, followed by evaluations of result consistency and robustness across repeated runs.

\subsubsection{Case Study}

\begin{figure}[pos=h]
    \centering    \includegraphics[width=0.8\linewidth]{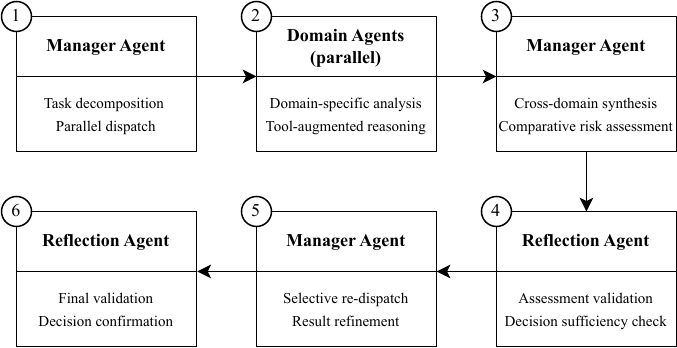}
    \caption{Workflow of AutoMARS case study for pairwise risk assessment.}
    \label{fig:case}
\end{figure}

We present a case study of a comprehensive pairwise risk assessment between Paulding County, GA (FIPS: 13223) and Cobb County, GA (FIPS: 13067) to illustrate the application of AutoMARS. While Figure~\ref{fig:inner-loop} shows the general multi-agent architecture, Figure~\ref{fig:case} depicts the concrete execution flow of this specific case, in which the workflow follows the revision branch and is routed back from Reflection Agent to Manager Agent for iterative refinement.

\paragraph{\textbf{Manager Agent: Initial Planning and Task Decomposition.}} At the beginning of this run, Manager Agent produced a concrete execution plan and issued four routing directives in parallel. The plan explicitly specified the required evidence types for each domain (e.g., socioeconomic indicators, hazard and loss metrics, air quality and facility-based exposure, and crash statistics) and defined a synthesis strategy to align risk direction and compare relative differences across counties. These domain-level comparisons were then combined based on three factors: magnitude, consistency, and reinforcement. Magnitude reflects how large the differences are, consistency captures whether multiple domains point to the same conclusion, and reinforcement considers whether risks amplify each other across domains. Domains with stronger and more consistent indicators, especially in high-impact areas, carry more weight in the final judgment.

\paragraph{\textbf{Domain Agents: Parallel Domain-Specific Analysis.}} Domain Agents execute in parallel to perform domain-specific analysis and generate structured evidence for comparative assessment. Each Domain Agent focuses on a distinct aspect of risk and applies tool-augmented reasoning by integrating authoritative datasets with LLM-based interpretation. All Domain Agents produce comparable, county-level evidence summaries that are aligned into a unified representation of directional risk indicators for subsequent cross-domain synthesis. This parallel execution ensures both efficiency and consistency in evidence generation across domains.

\textit{Census Agent.} Table~\ref{tab:census} highlights clear socioeconomic differences between two counties. Cobb County shows substantially higher overall exposure, reflected in its larger population scale and greater concentration of residents with multiple vulnerability factors. Its economic structure is more centralized, functioning as a regional employment hub with strong inward commuting patterns, whereas Paulding exhibits a smaller and more outward-oriented structure.

\begin{table}[pos=h]
    \centering
    \caption{Tools invoked by Census Agent and corresponding comparative socioeconomic indicators for Paulding and Cobb Counties.}
    \label{tab:census}
    \begin{tabular}{p{4cm} p{5.5cm} p{5.5cm}}
        \toprule
        \textbf{Tool} &
        \textbf{Paulding County} &
        \textbf{Cobb County} \\
        \midrule
        Socioeconomic Profile Tool (2019-2023) & Population: $\sim$174k; Labor Force: $\sim$90k; Median Household Income: \$94.6k; Unemployment Rate: 4.46\% & Population: $\sim$769k; Labor Force: $\sim$433k; Median Household Income: \$98.7k; Unemployment Rate: 4.55\% \\
        Community Resilience Tool (2023) & Population with 3+ Risk Factors: 12.13\% ($\sim$22k) & Population with 3+ Risk Factors: 14.96\% ($\sim$115k) \\
        Labor Exposure Tool (2023) & In-county Jobs: $\sim$29k; Commuting Pattern: Net outflow & In-county Jobs: $\sim$414k; Commuting Pattern: Net inflow \\
        \bottomrule
    \end{tabular}
\end{table}

\textit{Disaster Agent.} Table~\ref{tab:disaster} highlights key differences in hazard and disaster risk profiles between Paulding and Cobb Counties. Cobb County exhibits higher modeled multi-hazard risk and expected losses, along with lower community resilience, indicating a stronger baseline exposure to potential hazard impacts. In contrast, Paulding has experienced more frequent federally declared weather-related events in recent years, suggesting greater recent disaster occurrence. Observed severe weather impacts are broadly comparable in terms of economic losses, but differences emerge in event frequency and fatalities. Climate indicators further distinguish two counties, with Cobb showing conditions associated with greater heat-related exposure.

\begin{table}[pos=h]
    \centering
    \caption{Tools invoked by Disaster Agent and corresponding comparative hazard and disaster risk indicators for Paulding and Cobb Counties.}
    \begin{tabular}{p{4cm} p{5.5cm} p{5.5cm}}
        \toprule
        \textbf{Tool} &
        \textbf{Paulding County} &
        \textbf{Cobb County} \\
        \midrule
        Multi-Hazard Risk Index Tool & Composite Risk: 69.97; Expected Annual Loss: 79.52; Community Resilience: 44.12 & Composite Risk: 94.37; Expected Annual Loss: 95.85; Community Resilience: 28.72 \\

        Disaster Declaration Tool (2020-Present) & 6 declarations; includes Tropical Storm Zeta (2020), Hurricane Helene (2024), Winter Storm (2026) & 4 declarations; includes Hurricane Helene (2024), Winter Storm (2026) \\

        Severe Weather Tool (2023) & Events: 355; Damage: $\sim$\$37.4M; Deaths: 2; Injuries: 8 & Events: 319; Damage: $\sim$\$37.3M; Deaths: 1; Injuries: 8 \\

        Climate Tool (2023) & Annual TMIN: $\sim$10.2--10.6$^\circ$C; Precipitation: $\sim$1174--1179 mm & Annual TMIN: $\sim$11.9$^\circ$C; Precipitation: $\sim$1080--1416 mm \\

        \bottomrule
    \end{tabular}
    \label{tab:disaster}
\end{table}

\textit{Environment Agent.} Table~\ref{tab:environment} highlights clear differences in environmental exposure between Paulding and Cobb Counties. Cobb County shows observed air quality measurements, including PM2.5 and ozone levels, while no county-coded observations are available for Paulding during the study period. In addition, Cobb exhibits a substantially larger industrial and waste-related facility footprint, including more industrial plants, multiple landfills, and a major power plant, whereas Paulding has a more limited set of such facilities.

\begin{table}[pos=h]
    \centering
    \caption{Tools invoked by Environment Agent and corresponding comparative environmental exposure indicators for Paulding and Cobb Counties.}
    \begin{tabular}{p{4cm} p{5.5cm} p{5.5cm}}
        \toprule
        \textbf{Tool} &
        \textbf{Paulding County} &
        \textbf{Cobb County} \\
        \midrule

        Air Quality Tool (2023) & No county-coded AQS observations for PM2.5, Ozone, or CO & PM2.5: $\sim$10.82 $\mu g/m^3$; Ozone: $\sim$0.030 ppm \\

        Environmental Facility Tool (2023) &
        $\sim$7 industrial plants; 1 landfill; $\sim$1 chemical facility &
        $\sim$30 industrial plants; $\sim$6 landfills; $\sim$35 chemical facilities; includes a major natural gas power plant\\

        \bottomrule
    \end{tabular}
    \label{tab:environment}
\end{table}

\textit{Traffic Agent.} Table~\ref{tab:traffic} highlights clear differences in transportation safety risk between Paulding and Cobb Counties. Cobb County exhibits a substantially higher level of traffic activity, reflected in a much larger number of reported crashes. In addition, severe or fatal incidents are observed in Cobb but not in Paulding during the study period, indicating the presence of higher-impact events. While average severity levels are broadly similar, the occurrence of more severe incidents in Cobb points to a more elevated overall safety risk profile.

\begin{table}[pos=h]
    \centering
    \caption{Tools invoked by Traffic Agent and corresponding comparative transportation safety indicators for Paulding and Cobb Counties.}
    {
    \begin{tabular}{p{4cm} p{5.5cm} p{5.5cm}}
        \toprule
        \textbf{Indicator} &
        \textbf{Paulding County} &
        \textbf{Cobb County} \\
        \midrule
        Traffic Accident Tool (2023) &
        Total Crashes: 13; Severe/Fatal Crashes: 0 (0\%); Mean Severity: 2.00 &
        Total Crashes: 606; Severe/Fatal Crashes: 18 (3.0\%); Mean Severity: 2.06 \\
        \bottomrule
    \end{tabular}
    }
    \label{tab:traffic}
\end{table}

\paragraph{\textbf{Manager Agent: Cross-Domain Synthesis and Comparative Assessment.}} After receiving outputs from all Domain Agents, Manager Agent validates that the collected evidence is sufficient and consistent, and then performs a cross-domain synthesis integrating disaster, environmental, transportation, and socioeconomic indicators into a unified comparative assessment. Across domains, the evidence consistently indicates higher overall risk in Cobb County. Disaster-related indicators suggest higher modeled hazard intensity and expected losses accompanied by lower resilience. Environmental analysis reveals elevated particulate pollution levels and a denser concentration of industrial and waste-related sources, indicating greater potential for chronic exposure. Transportation indicators show substantially higher crash frequency and the presence of severe or fatal incidents. Socioeconomic evidence further indicates a larger and more vulnerable population, increasing the potential impact of hazards and environmental burdens. Based on the combined magnitude and consistency of these indicators across domains, Manager Agent determines that Cobb County represents the higher-risk location relative to Paulding County.

\paragraph{\textbf{Reflection Agent: Assessment Validation and Sufficiency Check.}} Upon receiving the integrated assessment, Reflection Agent evaluates whether the synthesized result is sufficiently supported for decision-making by examining domain coverage, structural completeness, and cross-domain consistency. During this process, minor inconsistencies are identified between modeled hazard risk metrics (e.g., NRI scores and expected annual loss) and interpretations based on disaster declarations and observed events. In addition, limitations are noted in the transportation safety analysis, which relies on a single-year snapshot and is not normalized by exposure factors such as population size or traffic volume. Although these issues do not invalidate the overall analytical framework, Reflection Agent determines that targeted refinement is necessary to improve the robustness of the supporting evidence. The workflow is therefore routed back to Manager Agent with guidance to resolve inconsistencies in hazard interpretation through numerical comparisons and reassess the role of transportation indicators while preserving the overall analytical structure.

\paragraph{\textbf{Manager Agent: Targeted Revision and Evidence Refinement.}} Following feedback from Reflection Agent, Manager Agent performs a targeted revision of the assessment while preserving previously validated evidence. In the disaster domain, categorical interpretations of FEMA National Risk Index (NRI) scores are removed, and the comparison is revised using only numerical indicators such as composite risk, expected annual loss, and community resilience. Although Paulding County reports more recent disaster declarations, Cobb County consistently exhibits higher numerical hazard risk indicators. In the traffic domain, Manager Agent acknowledges that crash statistics are based on a single-year snapshot and are not normalized by exposure factors such as population or traffic volume. As a result, transportation indicators are treated cautiously and excluded from primary decision weighting. The revised synthesis therefore places greater emphasis on more consistent evidence from socioeconomic, environmental, and hazard-related analyses, which collectively continue to indicate higher overall risk in Cobb County relative to Paulding County. The revised assessment is then forwarded to Reflection Agent for validation.

\begin{figure}[pos=h]
    \centering    \includegraphics[width=0.9\linewidth]{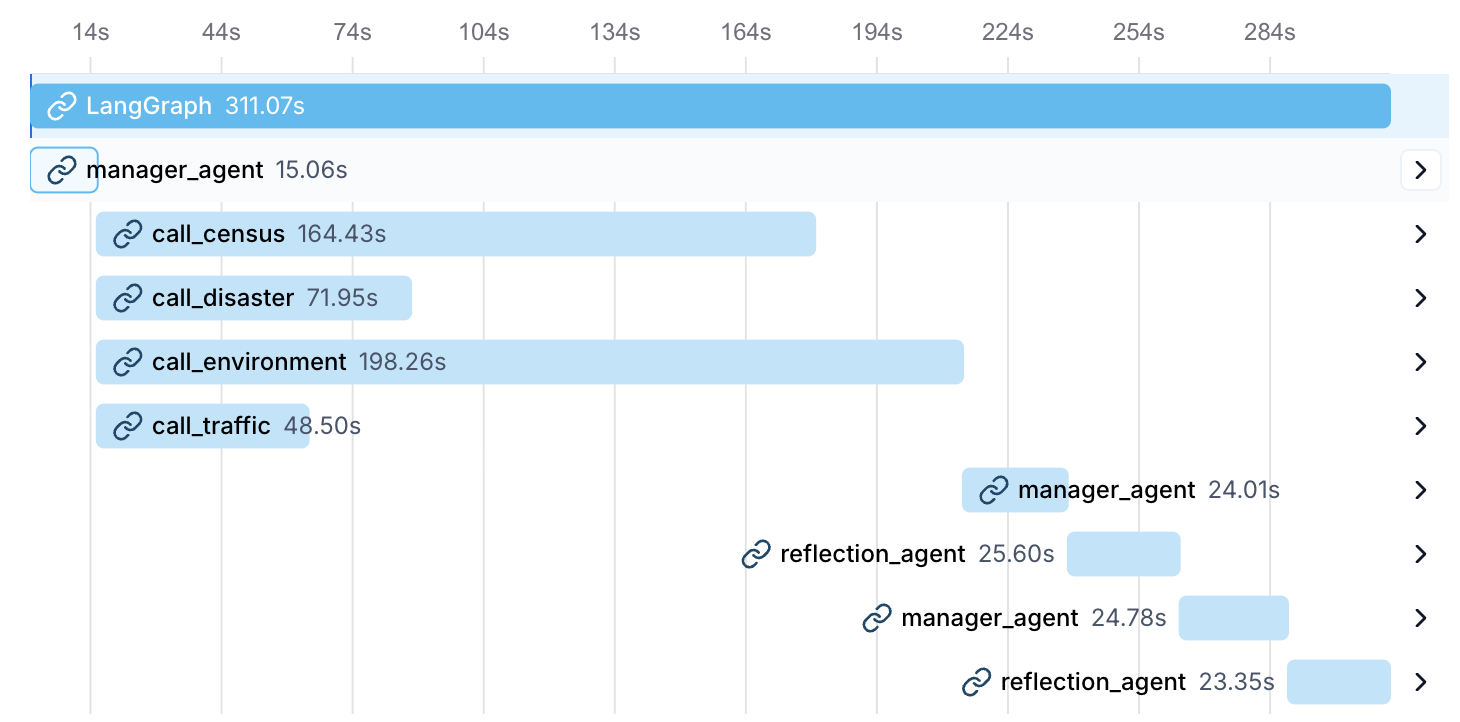}
    \caption{Execution timeline of AutoMARS for pairwise risk assessment.}
    \label{fig:waterfall}
\end{figure}

\paragraph{\textbf{Reflection Agent: Final Validation and Acceptance.}} After receiving the revised assessment, Reflection Agent conducts a final validation and confirms that the previously identified issues have been sufficiently addressed, including inconsistencies in hazard interpretation and limitations in transportation data. The revised synthesis is supported by consistent cross-domain evidence, with socioeconomic exposure, modeled disaster risk, and environmental burden providing mutually reinforcing indicators. Based on the convergence and relative strength of the remaining evidence, Reflection Agent accepts the assessment and confirms that Cobb County exhibits higher overall risk relative to Paulding County. This conclusion is primarily driven by higher modeled hazard risk and expected losses, greater chronic environmental burden, and a substantially larger and more vulnerable population in Cobb County. The workflow is then terminated, indicating that the assessment is sufficiently robust for decision-making. To provide a holistic view of the execution process, Figure~\ref{fig:waterfall} presents the overall execution timeline of AutoMARS workflow.

\subsubsection{Comparison Stability}

We evaluate the stability of the pairwise comparative assessment by conducting repeated experiments for each county pair. Specifically, for each ordered pair (A, B), we perform 20 independent runs to measure the consistency of the model’s judgment. To examine the potential impact of input order, we also reverse the pair and conduct an additional 20 runs for (B, A). This design enables us to assess both the robustness of the comparative outcomes across repeated runs and the sensitivity of the model to prompt ordering.

We select three representative county pairs with varying levels of risk difference to conduct this analysis, including Fulton vs Cobb, Gwinnett vs Clayton, and DeKalb vs Henry. The results are summarized in Table~\ref{tab:pairwise_stability}. We observe that the comparative outcomes are highly consistent across repeated runs. For all three pairs, the same county is identified as more risky in nearly all cases, with most comparisons achieving 100\% agreement and only minor variation observed in one pair. Importantly, reversing the input order does not change the overall conclusions, indicating that the model’s judgments are robust to prompt ordering. These findings demonstrate that the pairwise comparative framework produces stable and reliable risk assessments across repeated evaluations.

\begin{table}[pos=h]
    \centering
    \caption{Pairwise comparison stability across repeated runs.}
    {
    \begin{tabular}{llll}
        \toprule
        \textbf{County A} &
        \textbf{County B} &
        \textbf{More Risky} &
        \textbf{Percentage} \\
        \midrule
        Fulton & Cobb & Fulton & 100\% \\
        Cobb & Fulton & Fulton & 100\% \\
        Clayton & Gwinnett & Gwinnett & 100\% \\
        Gwinnett & Clayton & Gwinnett & 85\% \\
        DeKalb & Henry & DeKalb & 100\% \\
        Henry & DeKalb & DeKalb & 100\% \\
        \bottomrule
    \end{tabular}
    }
    \label{tab:pairwise_stability}
\end{table}

\subsection{Ranking Experimental Results}

We evaluate AutoMARS under three experimental scenarios corresponding to different spatial scales within Atlanta metropolitan area, including Atlanta Urban Core (5 counties), ARC region (11 counties), and Metro Atlanta (29 counties). In all experiments, GPT-5 is used as the backbone LLM to support risk assessment and cross-domain reasoning. For each scenario, county rankings are derived based on pairwise comparative assessments.

\subsubsection{Ranking Stability}

This part evaluates the stability of ranking results across repeated runs under different spatial scales. For visualization purposes, the runs in each figure are arranged using a quantitative criterion that minimizes the total distance between consecutive rankings, where distance is defined as the sum of absolute position changes across counties. This ordering improves visual continuity by reducing abrupt transitions between adjacent runs, while leaving the underlying experimental results unchanged.

\begin{figure*}[pos=h]
    \centering

    \begin{minipage}{0.48\textwidth}
        \centering

        \begin{subfigure}{\linewidth}
            \centering
            \includegraphics[width=0.8\linewidth]{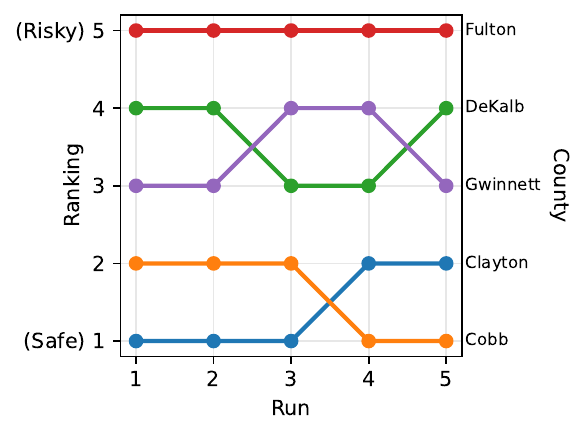}
            \caption{Atlanta Urban Core}
            \label{fig:ranking_stability_core}
        \end{subfigure}

        \begin{subfigure}{\linewidth}
            \centering
            \includegraphics[width=0.8\linewidth]{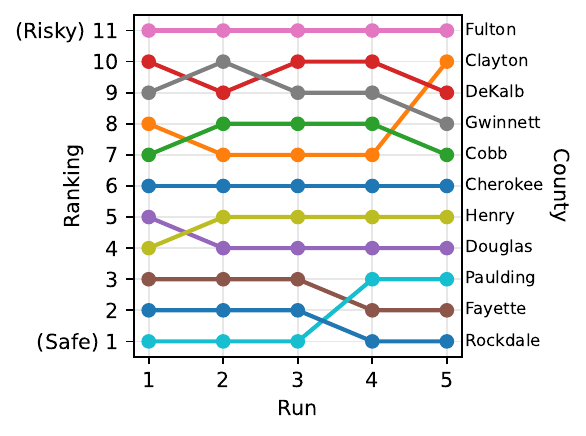}
            \caption{Atlanta Regional Commission}
            \label{fig:ranking_stability_arc}
        \end{subfigure}
    \end{minipage}
    \begin{minipage}{0.48\textwidth}
        \centering
        \begin{subfigure}{\linewidth}
            \centering
            \includegraphics[width=0.9\linewidth]{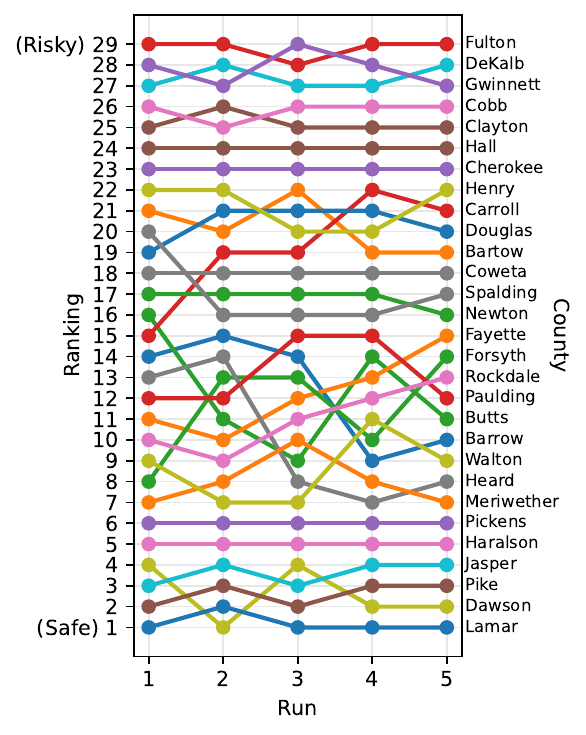}
            \caption{Metro Atlanta}
            \label{fig:ranking_stability_metro}
        \end{subfigure}
    \end{minipage}

    \caption{Ranking stability across repeated runs at different spatial scales.}
    \label{fig:ranking_stability}
\end{figure*}

For Atlanta Urban Core (Figure \ref{fig:ranking_stability_core}), rankings are highly stable across runs. Fulton County is consistently identified as the highest-risk location, while Clayton and Cobb Counties alternate as the lowest-risk locations. DeKalb and Gwinnett Counties remain in the middle positions with minor alternations, reflecting small variations between counties with relatively close risk levels. Overall, the ordering is highly consistent with only limited position swaps among intermediate counties.

For ARC region (Figure \ref{fig:ranking_stability_arc}), rankings remain largely stable as the number of counties increases. Most counties maintain consistent relative positions across runs, with only small variations observed. In particular, Clayton County exhibits slight fluctuations in its ranking, while the majority of other counties show stable ordering. This indicates that the framework preserves consistent comparative relationships even in moderately larger settings.

For Metro Atlanta region (Figure \ref{fig:ranking_stability_metro}), which represents the largest and most complex setting, rankings continue to exhibit strong stability at both extremes. The highest-risk and lowest-risk counties remain consistent across runs, while variations are primarily concentrated in the middle-ranked counties. These fluctuations can be attributed to the relatively small differences in risk levels among these counties, making them more difficult to distinguish consistently. Overall, the results demonstrate that AutoMARS produces robust and reliable ranking structures, with variability largely limited to cases where risk profiles are inherently similar.

\subsubsection{Ranking Statistics}

\begin{figure}[pos=h]
    \centering
    \begin{subfigure}[t]{0.32\textwidth}
        \centering
        \includegraphics[width=1\linewidth]{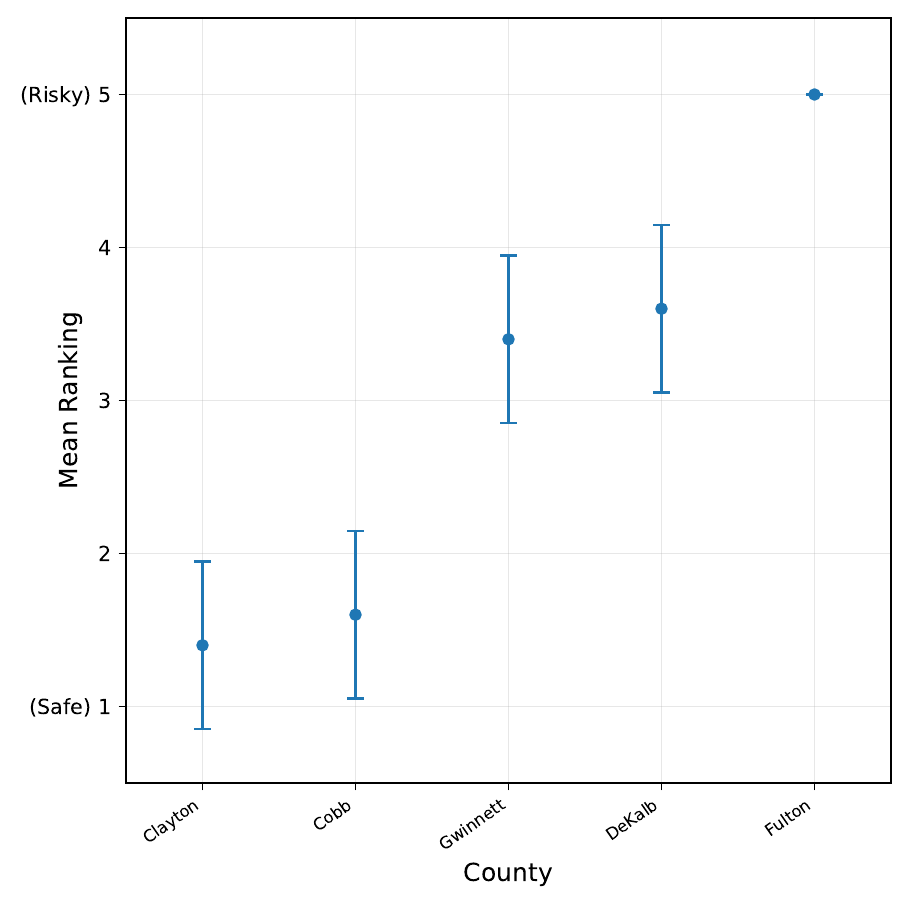}
        \caption{Atlanta Urban Core}
        \label{fig:mean_std_core}
    \end{subfigure}
    \hfill
    \begin{subfigure}[t]{0.32\textwidth}
        \centering
        \includegraphics[width=1\linewidth]{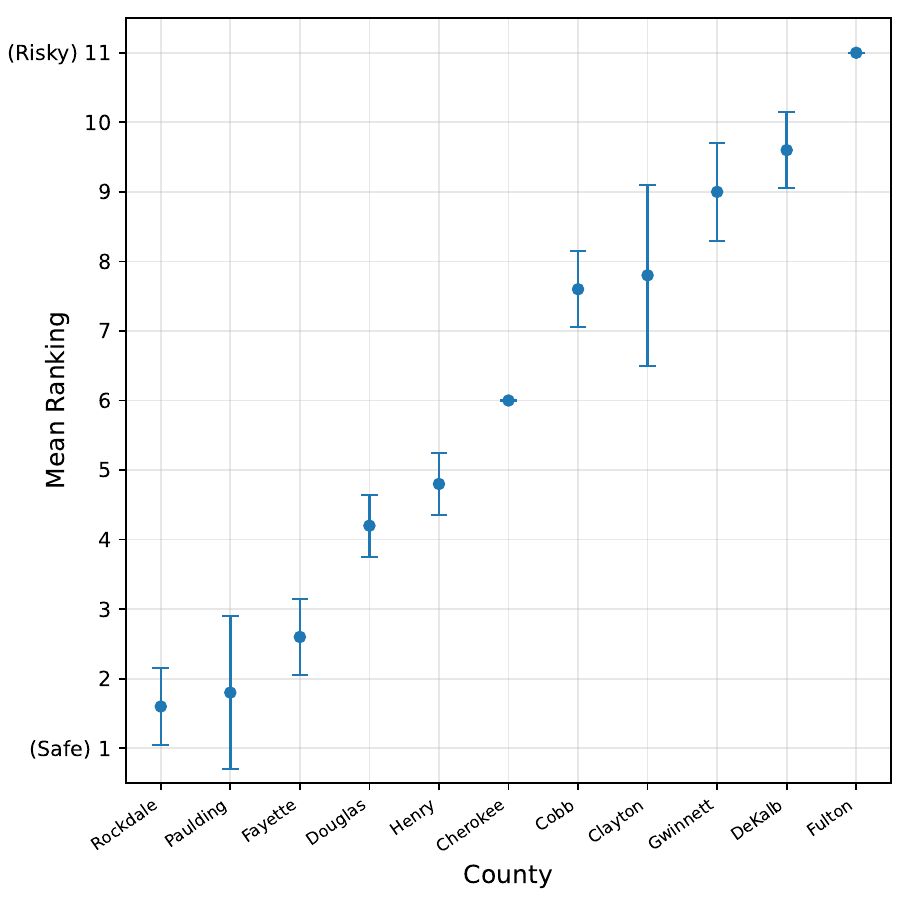}
        \caption{Atlanta Regional Commission}
        \label{fig:mean_std_arc}
    \end{subfigure}
    \hfill
    \begin{subfigure}[t]{0.32\textwidth}
        \centering
        \includegraphics[width=1\linewidth]{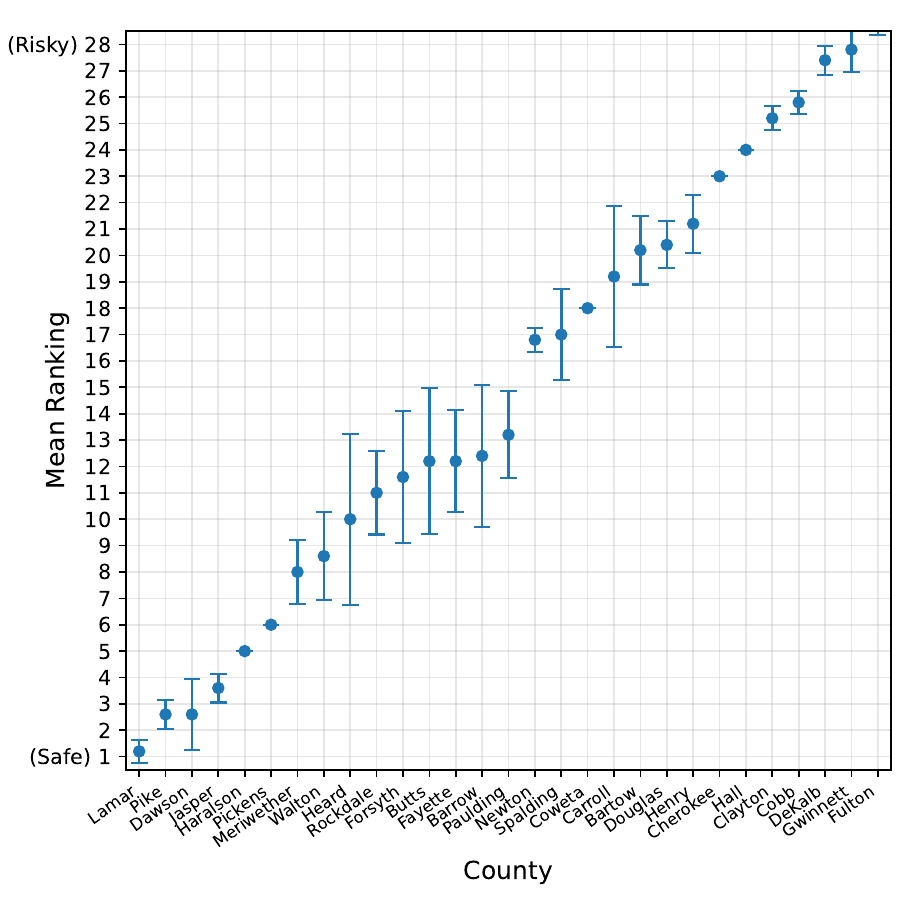}
        \caption{Metro Atlanta}
        \label{fig:mean_std_metro}
    \end{subfigure}

    \caption{Average and standard deviation of county rankings across repeated runs.}
    \label{fig:mean_std}
\end{figure}

We further analyze the statistical properties of the ranking results by examining the average ranking and standard deviation across repeated runs, as shown in Figure \ref{fig:mean_std}. For Atlanta Urban Core (Figure \ref{fig:mean_std_core}), the results exhibit clear and stable grouping patterns. Fulton County is consistently ranked as the highest-risk location with zero variance, indicating perfect stability. DeKalb and Gwinnett Counties form a middle group with similar average rankings and small variations, while Clayton and Cobb Counties are consistently ranked as the lowest-risk locations with closely aligned mean positions. Overall, the small standard deviations confirm that ranking variability is minimal in this setting.

For ARC region (Figure \ref{fig:mean_std_arc}), the ranking statistics remain highly stable as the number of counties increases. Most counties exhibit standard deviations within one ranking position, indicating strong consistency across runs. Slightly larger variations are observed for a small subset of counties, such as Paulding and Clayton, but the overall ranking structure remains largely unchanged. In Metro Atlanta region (Figure \ref{fig:mean_std_metro}), a more structured pattern emerges despite the increased complexity. Counties can be broadly grouped into four tiers: the highest-risk and lowest-risk groups both show very small standard deviations and highly stable rankings, while the middle counties form two subgroups. The first subgroup (e.g., Meriwether to Paulding) remains relatively stable in the lower-to-middle risk range, whereas the second subgroup (e.g., Newton to Henry) occupies the middle-to-higher risk range with slightly greater variation. These patterns are consistent with the qualitative observations from the ranking results presented earlier.

\begin{figure}[pos=h]
    \centering
    \begin{subfigure}[t]{0.32\textwidth}
        \centering
        \includegraphics[width=1\linewidth]{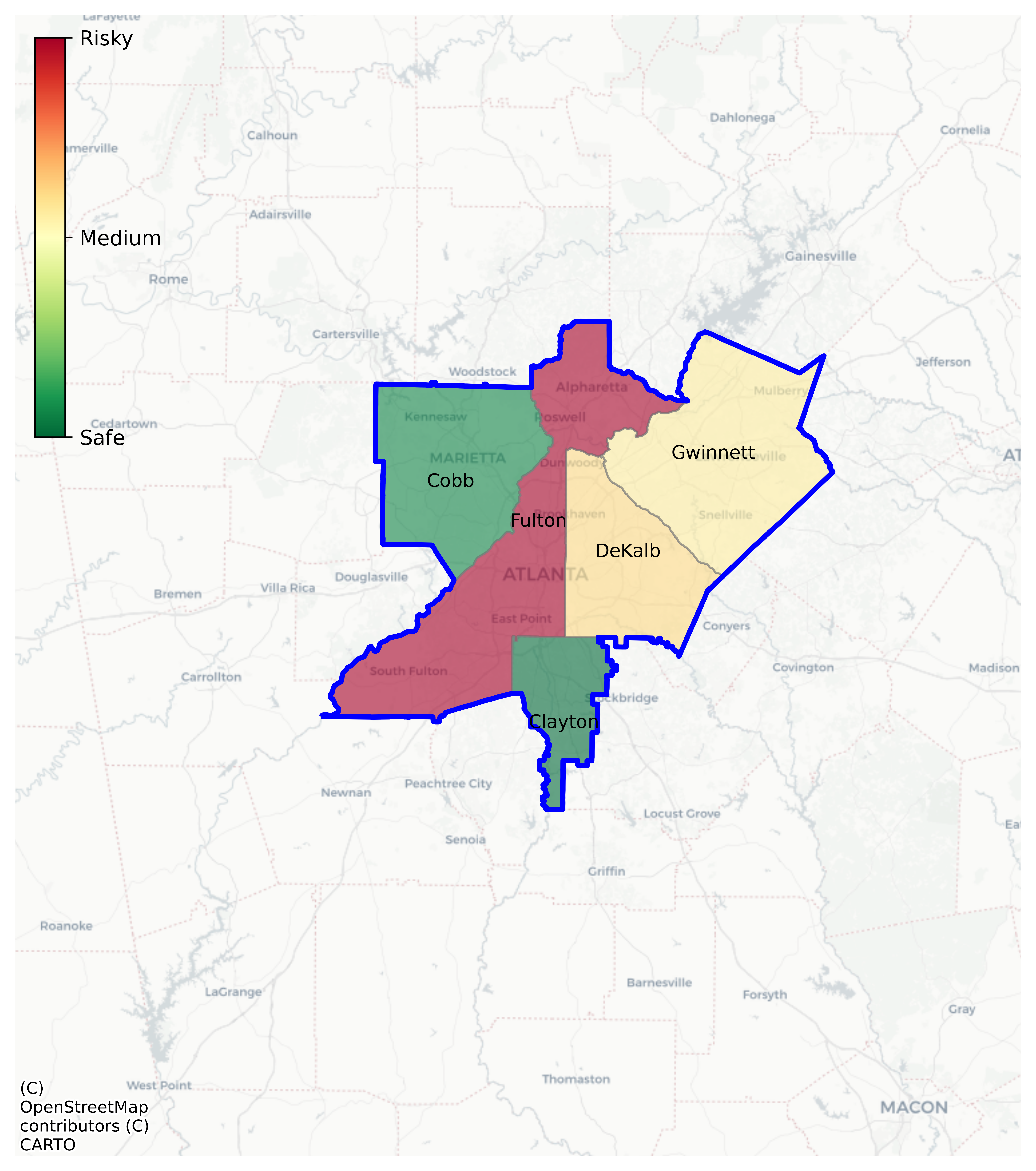}
        \caption{Atlanta Urban Core}
        \label{fig:map_core}
    \end{subfigure}
    \hfill
    \begin{subfigure}[t]{0.32\textwidth}
        \centering
        \includegraphics[width=1\linewidth]{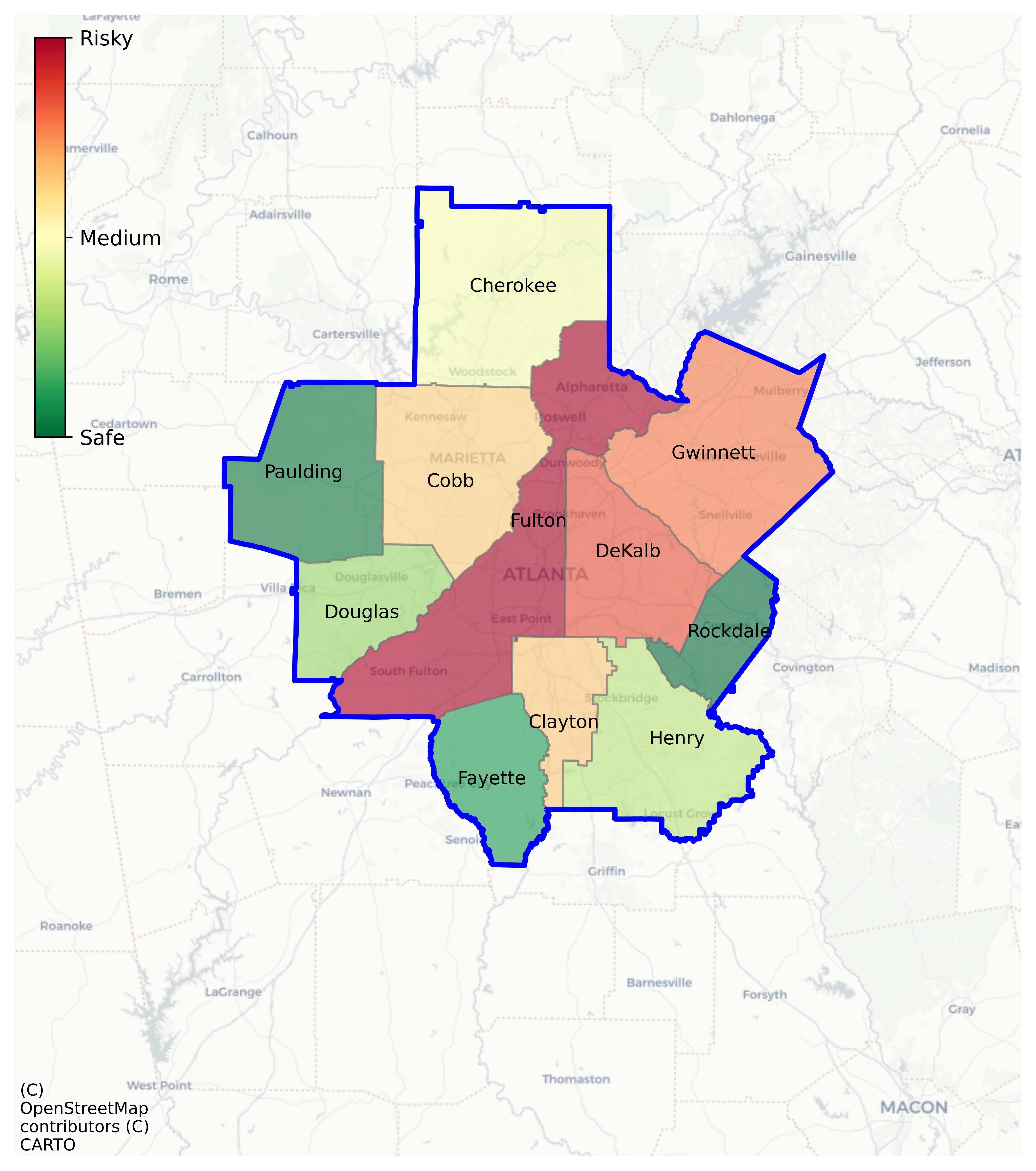}
        \caption{Atlanta Regional Commission}
        \label{fig:map_arc}
    \end{subfigure}
    \hfill
    \begin{subfigure}[t]{0.32\textwidth}
        \centering
        \includegraphics[width=1\linewidth]{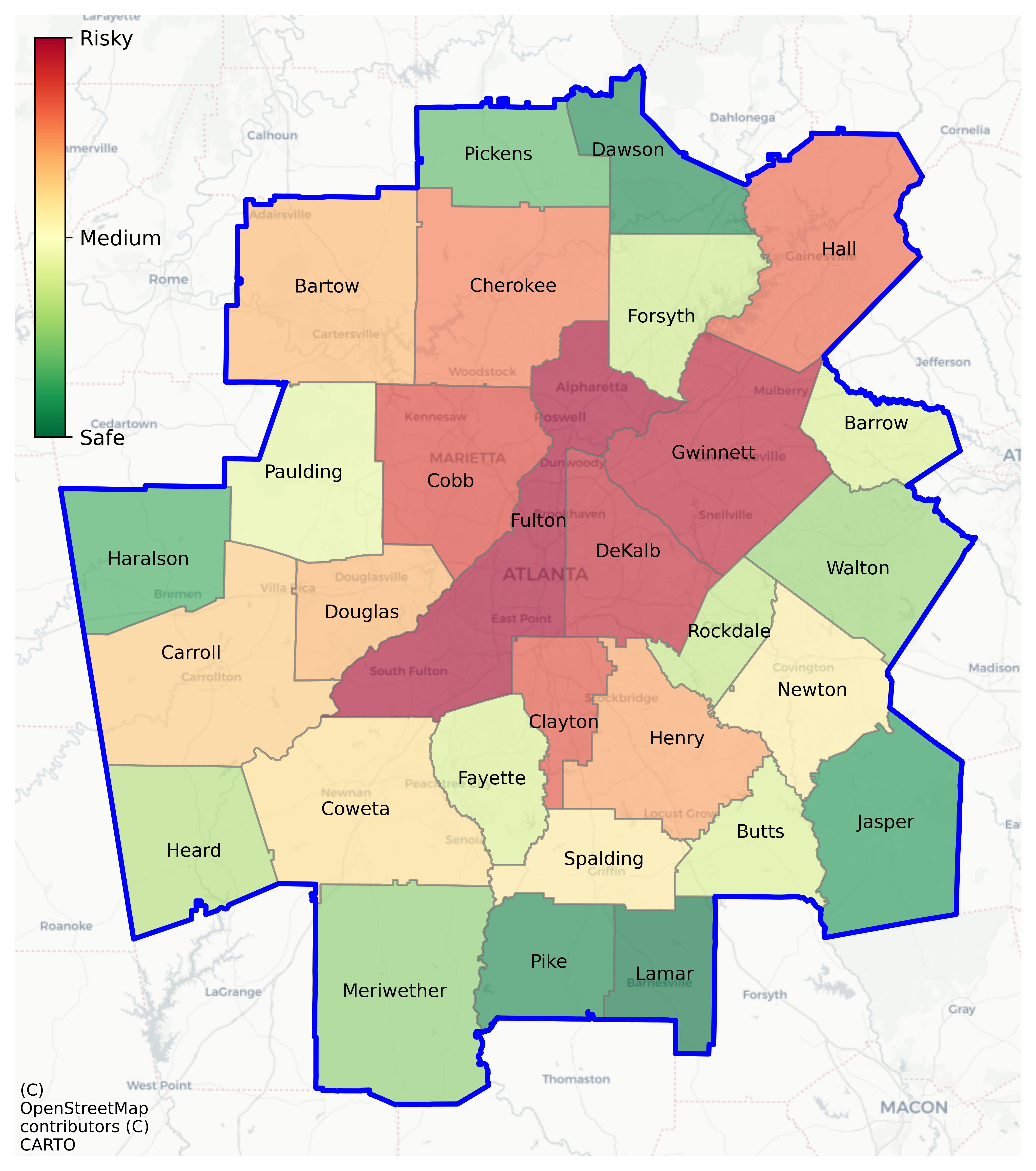}
        \caption{Metro Atlanta}
        \label{fig:map_metro}
    \end{subfigure}

    \caption{Spatial distribution of mean county rankings across Atlanta regions.}
    \label{fig:map}
\end{figure}

To provide a spatial interpretation of ranking outcomes, Figure \ref{fig:map} visualizes the mean rankings across regions. The maps highlight geographic distribution of relative risk levels, where color intensity reflects average ranking positions. This visualization provides an intuitive overview of how risk is distributed spatially, complementing the statistical analysis with a geographic perspective. Risk assessment in this paper is not derived from a single metric, but is based on a comprehensive integration of more than a dozen indicators spanning socioeconomic, disaster, environmental, and traffic domains. This multi-indicator approach captures diverse aspects of exposure, vulnerability, and hazard, which can lead to noticeable differences in risk levels even between adjacent counties.

\subsubsection{Model Configurations}

Table~\ref{tab:pipeline_performance} shows a highly consistent ranking pattern across five Atlanta Urban Core counties under different model configurations. Fulton consistently receives the highest risk ranking, followed by DeKalb and Gwinnett, while Cobb and especially Clayton tend to exhibit comparatively lower risk levels. Although minor ordering differences appear between configurations, the overall relative risk structure remains stable across all experiments. This indicates that the framework captures persistent regional characteristics rather than producing rankings that are overly sensitive to individual model choices. In addition, the standard deviations are generally small across counties and model settings, with many values below 1.0 and several rankings showing zero variance. The limited variability suggests that the framework produces robust and repeatable outputs even when replacing the underlying LLMs with GPT-5, GPT-5 mini\footnote{\url{https://developers.openai.com/api/docs/models/gpt-5-mini}}, or GPT-5 nano\footnote{\url{https://developers.openai.com/api/docs/models/gpt-5-nano}} variants. Overall, these results demonstrate that the multi-agent pipeline is relatively insensitive to model scale differences and maintains stable county-level prioritization across repeated runs.

\begin{table}[pos=h]
    \centering
    \caption{Performance comparison of end-to-end agent pipelines on Atlanta Urban Core with mean and standard deviation.}
    {
    \begin{tabular}{lllrrrrr}
        \toprule
        \textbf{Manager} & \textbf{Domain} & \textbf{Reflection} &
        \textbf{Clayton} &
        \textbf{Cobb} &
        \textbf{Gwinnett} &
        \textbf{DeKalb} &
        \textbf{Fulton} \\
        \midrule
        GPT-5 & GPT-5 & GPT-5 & $1.4 \pm 0.55$ & $1.6 \pm 0.55$ & $3.4 \pm 0.55$ & $3.6 \pm 0.55$ & $5.0 \pm 0.00$ \\
        GPT-5 mini & GPT-5 mini & GPT-5 mini & $1.8 \pm 0.84$ & $1.6 \pm 0.89$ & $2.6 \pm 0.55$ & $4.0 \pm 0.00$ & $5.0 \pm 0.00$ \\
        GPT-5 nano & GPT-5 nano & GPT-5 nano & $1.0 \pm 0.00$ & $2.4 \pm 0.55$ & $2.8 \pm 0.84$ & $3.8 \pm 0.45$ & $5.0 \pm 0.00$ \\
        \bottomrule
    \end{tabular}
    }
    \label{tab:pipeline_performance}
\end{table}

\subsubsection{Computational Efficiency}

Table~\ref{tab:parallel_runtime} presents the impact of parallel execution on the overall runtime of AutoMARS framework. Both sequential and parallel settings were evaluated over 10 runs, and the reported results correspond to mean runtime and standard deviation. The results demonstrate that parallel execution significantly improves computational efficiency compared with sequential execution. By allowing independent domain-specific agents to operate concurrently, the framework effectively reduces idle waiting time and improves overall resource utilization. In addition, the relatively stable standard deviations across repeated experiments indicate that the framework maintains consistent and reproducible runtime performance under both execution settings. Overall, these findings highlight the effectiveness of asynchronous multi-agent coordination in improving the scalability and responsiveness of the proposed AutoMARS framework.

\begin{table}[pos=h]
    \centering
    \caption{Impact of parallel execution on overall runtime with mean and standard deviation.}
    {
    \begin{tabular}{lr}
        \toprule
        \textbf{Setting} & \textbf{Total Runtime (s)} \\
        \midrule
        Sequential execution & $622.11 \pm 59.84$ \\
        Parallel execution & $327.49 \pm 64.70$ \\
        \bottomrule
    \end{tabular}
    }
    \label{tab:parallel_runtime}
\end{table}

Table~\ref{tab:tool_invoke_time} presents the impact of caching mechanism on tool invocation efficiency, using Air Quality Tool and Environmental Facility Tool as representative examples. For both caching and non-caching settings, the same query parameters were executed five times, and the reported values correspond to the average invocation time. The results show that caching substantially reduces repeated query latency for both tools. In particular, Air Quality Tool decreases from 76.68 seconds without caching to near-zero latency with caching, while the Environmental Facility Tool decreases from 37.98 seconds to 0.58 seconds on average. These results demonstrate that caching mechanism can effectively reduce redundant external API calls and significantly improve the computational efficiency of the framework.

\begin{table}[pos=h]
    \centering
    \caption{Impact of caching mechanism on tool invocation time.}
    {
    \begin{tabular}{lr}
        \toprule
        \textbf{Setting} & \textbf{Tool Invocation Time (s)} \\
        \midrule
        Air Quality Tool w/o caching & 76.68 \\
        Air Quality Tool w caching & 0.00 \\
        Environmental Facility Tool w/o caching & 37.98 \\
        Environmental Facility Tool w caching & 0.58 \\
        \bottomrule
    \end{tabular}
    }
    \label{tab:tool_invoke_time}
\end{table}

\section{Conclusion}
\label{sec:conlusion}

In this paper, we propose AutoMARS, an automatic multi-agent framework for scalable and comparative supply chain risk assessment. The framework enables end-to-end execution of analytical workflows by coordinating multiple specialized agents, integrating heterogeneous risk information across domains, and invoking external analytical tools. By combining structured task planning, multi-agent orchestration, and reflection-based validation, AutoMARS provides a systematic and interpretable approach to comparative risk evaluation. In addition, the framework supports parallel execution and a comparison-based sorting mechanism, enabling efficient analysis across multiple locations without relying on predefined scoring schemes.

Experimental results demonstrate effectiveness, robustness, and scalability of the proposed framework. Pairwise case study illustrates the ability of AutoMARS to synthesize cross-domain evidence and produces consistent comparative assessments. Stability analyses show that the framework generates reliable results across repeated runs, with strong consistency in both pairwise comparisons and ranking outcomes. Furthermore, ranking experiments across multiple spatial scales reveal that AutoMARS maintains stable ordering structures, with variability primarily limited to locations with similar risk profiles. Although adjacent counties appear geographically continuous, the observed red-green differences reflect underlying heterogeneity captured by more than a dozen indicators across socioeconomic, disaster, environmental, and traffic domains. Variations in population scale, vulnerability, economic structure, and exposure conditions can lead to substantial differences in risk even between neighboring counties. These boundaries therefore represent aggregated multi-dimensional differences rather than isolated factors. Nevertheless, because the analysis is conducted at the county level, neighboring counties may occasionally exhibit sharp local contrasts due to administrative boundary granularity and heterogeneous domain indicators. Decision-makers should therefore interpret localized ranking differences with appropriate caution.

Despite the strengths discussed above, some limitations remain. The current framework relies on manually designed prompts for LLM-based agents, making the results sensitive to prompt formulation and potentially affecting consistency across different runs and settings. Future work could address this limitation by exploring automatic prompt tuning techniques to improve the robustness and consistency of LLM-based agents. By adapting prompts based on data characteristics and task requirements, the framework can achieve more stable and reliable performance across different settings.

Overall, AutoMARS provides a unified and extensible framework for automated, multi-domain risk assessment in complex supply chain environments. By shifting from score-based evaluation to comparison-based reasoning, the approach improves the robustness and interpretability of risk assessment outcomes, particularly in settings involving LLM-based analytical processes. The proposed framework demonstrates the potential of multi-agent LLM systems to support scalable and structured decision-making, offering a practical foundation for risk-aware location evaluation in real-world supply chain applications.

\appendix

\bibliographystyle{cas-model2-names}

\bibliography{cas-refs}

\end{document}